\documentclass{article}
\usepackage{ijcai23}

\usepackage{times}
\usepackage{soul}
\usepackage{url}
\usepackage[hidelinks]{hyperref}
\usepackage[utf8]{inputenc}
\usepackage[small]{caption}
\usepackage{graphicx}
\usepackage{amsmath}
\usepackage{amsthm}
\usepackage{booktabs}
\usepackage{algorithm}
\usepackage{algorithmic}
\usepackage[switch]{lineno}

\usepackage{xcolor}
\newcommand{\red}[1]{#1}
\newcommand{\blue}[1]{#1}
\DeclareMathOperator*{\argmin}{\arg\min}
\usepackage{tabularx}
\usepackage{pifont}
\usepackage{multirow}
\usepackage{amsfonts}
\usepackage{bm}
\usepackage{enumitem}
\usepackage{subfigure}

\title{CTRL: Continuous-Time Representation Learning\\ on Temporal Heterogeneous Information Network}

\author{
Chenglin Li$^1$
\and
Yuanzhen Xie$^2$\and
Chenyun Yu$^{3}$\and
Lei Cheng$^2$\and
Bo Hu$^2$\and \\
Zang Li$^2$\and
Di Niu$^1$
\affiliations
$^1$University of Alberta, Edmonton, AB, Canada\\
$^2$Tencent, Shenzhen, Guangdong, China\\
$^3$Sun Yat-Sen University, Shenzhen, Guangdong, China
\emails
ch11@ualberta.ca,
xieyzh3@gmail.com
}
\begin{document}

\maketitle

\begin{abstract}
Inductive representation learning on temporal heterogeneous graphs is crucial for scalable deep learning on heterogeneous information networks (HINs) which \red{are} time-varying, such as citation networks.  
% \red{However, most existing approaches resort to taking snapshots of the continuous-time temporal graph, or are not inductive and thus cannot handle new nodes or edges. Most methods focus more on monographs and seldom pay attention to heterogeneous graphs.} 
% Furthermore, previous temporal graph embedding methods are often trained with either temporal node classification or link prediction tasks to simulate the \red{formulation} of temporal graphs, while ignoring the evolution of high-order topological structures on temporal graphs.
\red{However, most existing approaches are not inductive and thus cannot handle new nodes or edges. Moreover, previous temporal graph embedding methods are often trained with the temporal link prediction task to simulate the link formation process of temporal graphs, while ignoring the evolution of high-order topological structures on temporal graphs.}
% Moreover, previous methods model the evolution of a temporal network as a link formation process, thus, they often adopt the temporal link prediction task to train a model. 
To fill these gaps, we propose a Continuous-Time Representation Learning (CTRL) model on temporal HINs. 
To preserve heterogeneous node features and temporal structures, CTRL integrates three parts in a single layer, they are 1) a \emph{heterogeneous attention} unit that measures the semantic correlation between nodes, 2) a \emph{edge-based Hawkes process} to capture temporal influence between heterogeneous nodes, and 3) \emph{dynamic centrality} that indicates the dynamic importance of a node. 
We train the CTRL model with a future event (a subgraph) prediction task to capture the evolution of the high-order network structure.  
Extensive experiments have been conducted on three benchmark datasets. The results demonstrate that our model significantly boosts performance and outperforms various state-of-the-art approaches. Ablation studies are conducted to demonstrate the effectiveness of the model design. 
\end{abstract}

\section{Introduction}
Representation learning on graphs is gaining popularity due to the widespread presence of graph-structured data in real-world scenarios, e.g., citation networks and social networks. 
Most of these real-world graphs exhibit heterogeneity and complex dynamics that evolve continuously in time. 
\red{For example, a citation network may have the ``Author'' and ``Paper'' nodes, while an author can write multiple papers at different points in time.} 
% For example, a citation network may involve different types of nodes and edges, e.g., the ``Author'', ``Paper'',  and ``Venue''(or conference) nodes and citation edges between papers. Moreover, these nodes and edges may appear at different timestamps. 
Although an increasing amount of efforts have been made to study heterogeneous information networks (HINs) and temporal networks, most recent studies focus on either static HINs~\cite{wang2021self,wang2019heterogeneous} or temporal monographs~\cite{tgat_iclr20,wang2021inductive}. 
% HAN~\cite{wang2019heterogeneous}, 
% HeCo~\cite{wang2021self}

% 【current temporal HIN method and drawbacks】
% Cons 1: C1
A few recent studies attempt to handle the dynamics and heterogeneity of temporal HINs simultaneously. 
THINE~\cite{huang2021temporal} and HPGE~\cite{ji2021dynamic} leverage the attention mechanism and Hawkes process to handle the heterogeneity and dynamics in temporal HINs.
% capture temporal influence between heterogeneous nodes. 
% THINE~\cite{huang2021temporal} leverages the attention mechanism and meta-path to handle the heterogeneity and use the Hawkes process to model the evolution of temporal networks. 
% HPGE~\cite{ji2021dynamic} also leverages the Hawkes process and time-importance sampling techniques to model the dynamics of the temporal graph. 
% C1
% However, the output of these methods are whole-time node embeddings, i.e., a node has a single embedding at all timestamps~\cite{barros2021survey}, 
However, they produce the all-time general node embeddings, i.e., each node is assigned with a single embedding for all timestamps~\cite{barros2021survey}, and thus can not generate dynamic node embeddings at any given time. 
In addition, they are all transductive models that are not inductive and scalable to new nodes. 
\red{Therefore, there still exists a gap in the literature to develop deep inductive models for temporal HINs that can generate time-varying node embeddings in continuous time.}

% c2,
There are two significant challenges to obtaining inductive temporal HIN embeddings. The first challenge lies in better modelling the dynamic impact between heterogeneous nodes. 
% TREND~\cite{wen2022trend} integrates a single Hawkes process into the GNN layer to model the temporal influence between all node pairs in a network in an inductive manner. However, it can not handle the node heterogeneity in HINs. THINE~\shortcite{huang2021temporal} and HPGE~\shortcite{ji2021dynamic} create a node-specific Hawkes process for every node to model the dynamic influence or the exciting effects between nodes. 
% Previous works adopt either a single, e.g., TREND~\cite{wen2022trend}, or node-wise, e.g., THINE~\shortcite{huang2021temporal} and HPGE~\shortcite{ji2021dynamic}, Hawkes process to model the temporal influence between nodes. 
TREND~\cite{wen2022trend} integrates a single Hawkes process to model the temporal influence between nodes. However, it can not handle the node heterogeneity in HINs. THINE~\shortcite{huang2021temporal} and HPGE~\shortcite{ji2021dynamic} create a per-node Hawkes process to model the exciting effects between heterogeneous nodes. However, they are not inductive to new nodes and also fail to incorporate the heterogeneity of edges between heterogeneous nodes. 

% There are two serious challenges for \red{inductive} temporal HIN embedding. First, \emph{how to better model the dynamic impact between heterogeneous nodes?} 
% TREND~\cite{wen2022trend} integrates a single Hawkes process into the GNN layer to model the temporal influence between all node pairs in a network in an inductive manner. However, it can not handle the node heterogeneity in HINs. 
% THINE~\shortcite{huang2021temporal} and HPGE~\shortcite{ji2021dynamic} create node-specific Hawkes process for every node to model the dynamic influence or the exciting effects between nodes which is not inductive to new nodes and also ignores the heterogeneity of edges between nodes. 

% \red{Moreover, they only consider }

% c3 
% The other challenge is on \emph{how to get node embeddings that effectively capture the high-order local topological dynamics of temporal HINs?} 
The other challenge is how to effectively capture the evolution of network structures of temporal HINs. 
% Usually the dynamics of a graph include the addition or deletion of nodes and edges during its evolution. 
Capturing the network structure evolution is crucial to a better understanding of temporal HINs. 
Most existing works model the evolution of temporal networks as a link formation process where each temporal link/edge is treated as the basic evolution unit in the graph. 
Thus, they often adopt the temporal link prediction task to train the model, e.g., TGAT~\shortcite{tgat_iclr20}, CAW~\shortcite{wang2021inductive}, and TREND~\shortcite{wen2022trend}. 
\red{
% However, real-world temporal HINs often involve the addition of events with complicated high-order subgraph structures.
However, the evolution of real-world temporal HINs often involves the formation of events with complicated high-order subgraph structures. 
For example, the basic evolution unit in a citation network should be a \emph{publication of a paper}, which is a subgraph event including the authors of the paper, the venue the paper is published in, and the cited papers, where all the edges within the event are formed at the same time. 
% For example, the basic evolution unit in a citation network should be a publication of a paper, which is a subgraph event including the authors of the paper, the venue the paper is published, and the cited papers, where all the edges in the event happen at the same time.
% Capturing the occurrence possibility of such a basic event is critical in understanding the evolution of temporal HINs.
Capturing the formation process of such subgraph events with high-order local structures is critical to understanding the evolution of temporal HINs.
} 

% Most existing methods adopt multiple GNN layers to preserve structural information of the network and train the model via temporal link prediction or node classification task to further incorporate structure dynamics in node embedding. However, most of the    

% c4  --> little trick? Refers to the graphormer paper   
% Furthermore, 
\blue{To address the aforementioned challenges, in this paper, we propose CTRL, a novel Continue-Time Representation Learning model to capture the temporal influence between heterogeneous nodes and high-order structural evolution of temporal HINs. 
% We build node type-dependent feedforward layers to map the features of different types of nodes into the same latent space. Moreover, we build edge type-dependent matrices to further transform node representations during the message-passing process to handle edge heterogeneity. 
% Furthermore, three key factors, namely, semantic correlation, temporal influence, and dynamic centrality, are considered to determine the relative importance of neighbour nodes in the local aggregation process. 
% Specifically, the temporal influence which is modelled by an edge-based Hawkes process and dynamic centrality capture the dynamic importance of nodes.
The CTRL model incorporates temporal and dynamic structural information of temporal HINs into the Transformer framework~\cite{vaswani2017attention}. 
% Specifically, temporal information is represented by the temporal influence between heterogeneous nodes which is modelled by an edge-based Hawkes process. Then, the dynamic structural information is reflected by the dynamic centrality of graphs.
Specifically, temporal information which is modelled by an edge-based Hawkes process and dynamic structural information which is reflected by the dynamic centrality of graphs are incorporated into the aggregation stage of a CTRL layer. 
We train the model by predicting the event formation in temporal HINs. 
% Finally, we train the model by predicting the probability of the occurrence of basic events in temporal HINs. 
In proposing CTRL, we make the following contributions: }

% into the \red{graph attention/transformer framework}. 
% First, we explore an inductive graph embedding model to solve the problem of continue-time representation learning on temporal HINs.

% First, we propose a novel temporal graph neural network that aggregates the messages from temporal neighbours based on the three factors.
% They are 1) \emph{semantic correlation} between the target node and neighbours which are measured by the attention scores like previous static methods, e.g., GAT~\cite{velickovic2018graph}.
% First, we propose a novel temporal graph neural network that aggregates the messages from temporal neighbours based on the three factors. 

\red{
First, apart from the \emph{semantic correlation} between nodes which are measured by attention scores, we further consider two other key factors in the aggregation process of a CTRL layer. They are 1) \emph{temporal influence} between the target and neighbouring nodes and 2) \emph{dynamic centrality}, which measures the importance of a node over time. 
To be specific, we propose an edge-based Hawkes process where a neural network is used to extract edge-specific decay rates to capture temporal influence between heterogeneous nodes.
We further use the dynamic degrees of neighbours to weigh the importance of each neighbour to incorporate the dynamic centrality in the aggregation procedure. 
% Intuitively speaking, historical neighbours that are closer in time should be more influential. Thus, we propose an edge-based Hawkes process where an MLP module is built to generate an edge-specific decay rate to model the dynamic impact between heterogeneous nodes based on their representations. 
% Dynamic centrality is also considered to better understand a temporal network. Therefore, we further adopt the dynamic degrees of neighbours to weigh the importance of each neighbour in the aggregation process.
% Furthermore, the dynamic node degree is also a dynamic node feature, which is encoded into a real-valued vector and is added to the original node feature. 
In addition, the messages from heterogeneous temporal neighbours are transformed through node-type- and edge-type-dependent modules before the aggregation to handle graph heterogeneity.}

% We leverage the dynamic degree centrality from two perspectives.
% Specifically, we first encode the node degree into a real-valued vector which is added to the original node feature. Then, the dynamic degrees of neighbours are also used to weigh the importance of each neighbour in the aggregation process. 
% we map the dynamic node degree into the real-valued vector and add to the node feature  
% where an MLP module  to generate edge-specific  

% Secondly, to capture the high-order evolution patterns of temporal HINs, we propose to train the proposed model by predicting the occurrence probability of basic events. 
% Specifically, we average all node embeddings in an event (subgraph) and use an MLP module to predict the event probability. Furthermore, to preserve the structure information within an event, we build another MLP layer which takes the information of an edge (including the representations of the source and target node) as input and outputs the occurrence probability of the edge. 
Secondly, we propose to train CTRL through the prediction of temporal events to capture the evolution of high-order structures in temporal HINs. 
Specifically, two MLP sub-modules are built to predict the event occurrence probability and the probabilities of occurrence of all the edges within the event to preserve the dynamics of both the high-order and first-order local structure. 
% Specifically, we average all node embeddings in an event and use an MLP module to predict the event probability.  Moreover, to preserve the structural information within events, we further predict the probability of the formation of each edge in an event by feeding the latent representations of its source and target node into another MLP layer.
% for each edge in an event, the latent representations of its source and target node are fed into another MLP layer to predict the edge probability.  
By minimizing the prediction loss events and edges, the CTRL captures the evolution of the high-order local structure of temporal HINs.
% we build another MLP layer that takes the latent representations of both the source and target node of an edge as input and outputs the occurrence probability of the edge.
% Moreover, to preserve the structure information within events, we build another MLP layer that takes the information of an edge (including the representations of the source and target node) as input and outputs the occurrence probability of the edge.

We conducted extensive experiments on three public datasets to demonstrate the superiority of CTRL over a range of state-of-the-art approaches and baselines on the inductive temporal link prediction task. Moreover, we demonstrate the effectiveness of the model design choices through ablation studies. 

\section{Related Work}
% \noindent \textbf{Graph representation learning}
Graph embedding tries to represent nodes in a low-dimensional space while preserving node features and the topological structures of the graph~\cite{cai2018comprehensive,wu2020comprehensive}. Traditional methods mainly focus on the representation learning on static homogeneous networks, e.g., Deepwalk~\shortcite{perozzi2014deepwalk}, LINE~\cite{tang2015line}, GCN~\cite{kipf2017semi}, GAT~\cite{velickovic2018graph}, and do not consider the evolution of graphs.   

Recent graph embedding studies focus on two more practical scenarios, i.e., the heterogeneous information network (HIN) and temporal networks. 
For HIN, traditional shallow methods model semantics and structures based on meta-path~\cite{sun2012mining}, e.g., Meta-path2vec~\cite{dong2017metapath2vec} and HAN~\cite{wang2019heterogeneous}, while deep models incorporate heterogeneous node/edge information into the graph neural network through different techniques such as attention mechanism HetGNN~\cite{zhang2019heterogeneous} and HGT~\cite{hgt}. 
For temporal networks, most of the existing works resort to taking snapshots of the continuous-time temporal networks, and model the dynamics of node embeddings from a sequence of snapshots, e.g., DySAT~\cite{sankar2020dysat}, MTSN~\cite{liu2021motif}, and EvolveGCN~\cite{pareja2020evolvegcn}. Recent work focus more on modelling the evolution of networks in a continuous-time manner, e.g., TGAT~\cite{tgat_iclr20}, TREND~\cite{wen2022trend}, CAW~\cite{wang2021inductive}, and HVGNN~\cite{sun2021hyperbolic}. 
Besides, some work focuses on exploring the evolution patterns of temporal graphs such as the triadic closure process \cite{zhou2018dynamic} in social networks. 
There are also some works that aim at providing real-time node embedding services from temporal networks, e.g., APAN~\cite{wang2021apan}, TGL~\cite{zhou2022tgl} and TGN~\cite{rossi2020temporal}.

There is an increasing trend that focuses on temporal HIN embedding. Most existing works use meta-path to capture semantics and structures in HIN and take snapshots of the temporal network to model the dynamics of node embeddings, like DHNE~\cite{yin2019dhne}, Change2vec~\cite{bian2019network}, and DyHNE~\cite{wang2020dynamic}. 
Other than taking snapshots, 
HDGAN~\cite{li2020heterogeneous} leverages time-level attention to model network evolution. 
% HDGNN uses semi-supervised for dynamic HIN. 
Besides, some work takes advantage of the Hawkes process~\cite{hawkes1971spectra} to simulate the evolution of the temporal network, e.g., THINE~\cite{huang2021temporal} and HPGE~\cite{ji2021dynamic} for HIN, and HTNE~\cite{zuo2018embedding}, MMDNE~\cite{lu2019temporal}, and TREND~\cite{wen2022trend} for monograph. 
However, these methods are not inductive to new nodes, thus, there is still a lack of deep methods for temporal HINs. 
TGSRec~\cite{fan2021continuous} applied the TGAT model to the sequential recommendation problem but does not explicitly consider the influence of node heterogeneity. 
\red{To fill this gap, in this paper, we propose a deep graph neural network for continuous-time representation learning on temporal HINs.}

\begin{figure*}[ht]
  \centering
  \includegraphics[width=6in]{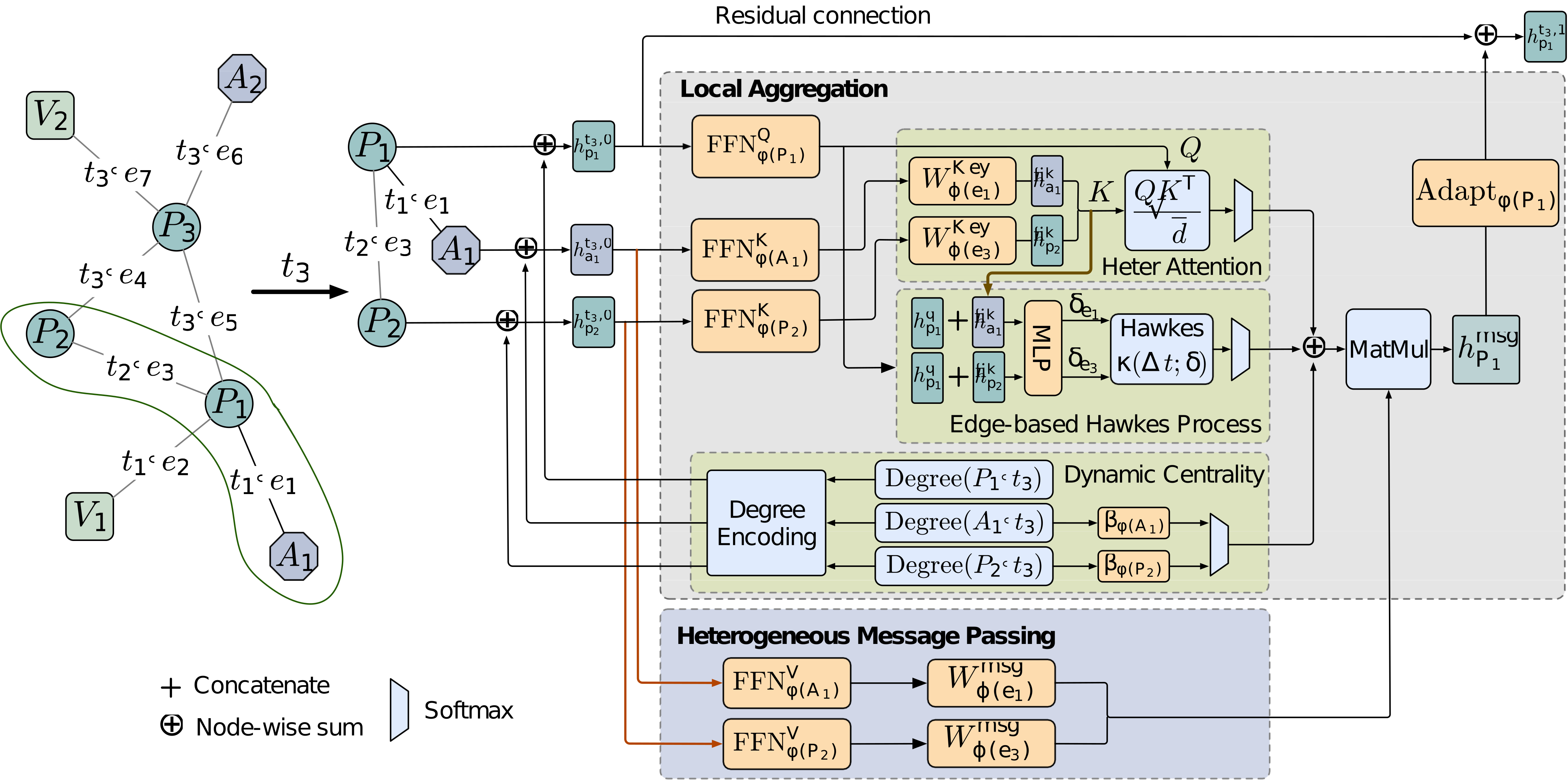}
  \caption{The architecture of a layer of the CTRL model. 
}
  \label{fig:sys}
\end{figure*}

\section{Preliminaries}
In this section, we recap the preliminaries in temporal HINs, graph neural network and Hawkes process. 

\noindent \textbf{Temporal HIN.}
% What is THIN and the goal of CTRL on THIN
Temporal HIN is defined as a graph $\mathcal{G} = (\mathcal{V}, \mathcal{E}, \mathcal{T}; \phi, \varphi)$ where $\mathcal{V}$ is the set of nodes, $\mathcal{E}$ indicates the set of edges and $\mathcal{T}$ represents the set of timestamps. 
$\phi: \mathcal{V} \mapsto \mathcal{A}$ and $\varphi: \mathcal{E} \mapsto \mathcal{R}$ are two mapping functions that map nodes and edges to their corresponding types. $\mathcal{A}, \mathcal{R}$ are the sets of types for nodes and edges, respectively. For HINs, $|\mathcal{A}| + |\mathcal{R}| > 2$. 

% Previous methods often define an edge $e^t_{ij} \in \mathcal{E}$, established between nodes $v_i$ and $v_j$ at time $t$, as the basic event during the evolution of a temporal HIN. However, in some networks, the basic event is a complicated subgraph that involves multiple nodes and edges. For example, in the citation network, the basic event is the publication of a paper that contains the authors of the paper, the venue where the paper is published, and the cited papers. 

% \noindent \textbf{Basic-event}

\noindent \textbf{Continuous-time Temporal HIN Embedding.}
Given a temporal HIN $\mathcal{G}$, the goal of continuous-time representation learning on temporal HIN is to learn an embedding function $f: \mathcal{V} \times \mathcal{T} \mapsto	\mathbb{R}^d \times \mathbb{T}$ where $d$ is the dimensionality of node embedding. Thus, we can get the low-dimensional representation of a node at any given time $t \in \mathbb{T}$. 

% What is Hawkes's process? how do we use it? (difference?) 

\noindent \textbf{Graph Neural Network.} The general architecture of a graph neural network (GNN) layer consists of two parts. They are \emph{Message passing}, which extracts messages from source nodes and passes them to the target node; and \emph{Aggregation}, which aggregates the messages from source nodes to update the representation of the target node.

In this work, we handle the network heterogeneity through node type- and edge-type-dependent modules in the message-passing stage and capture the node feature and temporal topological of a network in the aggregation stage.

\noindent \textbf{Hawkes Process.} 
Hawkes process~\cite{hawkes1971spectra} is a point process that assumes the arrival of an event is influenced by historical events.
It models the occurrence of an event $e$ at time $t$ by a conditional intensity function
\begin{equation} \label{eq:hawkes_org}
\lambda(e, t) = \mu(e, t) + \sum_{e_h,t_h; t_h < t} \kappa(t-t_h; \delta), 
\end{equation}
where $\mu(e, t)$ is the base intensity of event $e$ at time $t$, $\kappa(t-t_h, \delta) = exp(-\delta (t-t_h))$ denotes the time decay effect of historical event $e_h$ and $\delta$ is the decay rate. 
% Previous works for monograph use a single Hawkes process, i.e., a single $\delta$, to model the influence between all the node pairs, e.g., TREND~\shortcite{wen2022trend}. \red{However, in HINs, the } 
% In HIN, 

In this work, we apply the time decay effect to determine the temporal importance of neighbour nodes in the aggregation process of a GNN layer which is equivalent to the conditional intensity in~\eqref{eq:hawkes_org},  as demonstrated in TREND~\shortcite{wen2022trend}.

\section{Methodology}
\label{sec:sys}

In this section, we present the proposed CTRL for temporal HINs. First, we elaborate on the key designs on how we capture the temporal evolution of node features and network structure. Then, we introduce the future event prediction task for the optimization of the proposed GNN model. 

\subsection{Model Overview}
\label{model_overview}
Figure~\ref{fig:sys} shows the overall structure of a single layer of the CTRL model. 
\red{An example is given on how a CTRL layer extracts the temporal representation of node $P_1$ at time $t_3$. 
To be specific, 
we first sample its historical neighbors $(A_1,P_2)$ which are connected with $P_1$ through edges $t_1, e_1$ and $t_2, e_3$, respectively.
% At time $t_3$, given a target node $P_1$ and its temporal neighborhood $(A_1,P_2)$ at $t_1$ and $t_2$, respectively. The GNN layer takes their hidden representations (or features), timestamps, and dynamic node degrees as input and outputs the time-aware representation for the target node at the given time. We denote the hidden representation of node $P_1$ at time $t_3$ from the l-th layer as $h^{t_3, l}_{P_1}$. 
Then, the encoded dynamic centrality, indicating the dynamic importance of a node, is added to the node features which are the input of the first CTRL layer, i.e., $h^{t,0}_{P_1}$, $h^{t,0}_{A_1}$, and $h^{t,0}_{P_2}$. 
% Note that we leverage the dynamic degree centrality, which is defined as the number of links incident upon a node before a given timestamp, to measure the importance of neighbour nodes of a target node. 
In message passing, the representations of neighbours are passed to the target node through node type- and edge-type dependent modules, i.e., $\text{FFN}^V_{\phi(\cdot)}$ and $W^\text{msg}_{\varphi(\cdot)}$. 
In addition, we consider three key factors to weigh the importance of neighbour nodes during the local aggregation process. 
They are 1) \emph{semantic correlation} which is calculated via the heterogeneous attention module, 
% between the representations of target and neighbourhood nodes; 
2) \emph{temporal influence} which is measured by the proposed edge-based Hawkes process, 
3) and \emph{dynamic centrality} measured by the dynamic degree centrality, which is defined as the number of links incident upon a node before a given timestamp. 
% Note that we leverage the dynamic degree centrality, which is defined as the number of links incident upon a node before a given timestamp, to measure the importance of neighbour nodes of a target node.
}
A node type-dependent adapt module, i.e., $\text{Adapt}_{\phi(P_1)}$, is built to transform the aggregated message, $\bm{h}_m$, into the feature distribution of the target node. Finally, the adapted message is added to the representation of the target node from the last layer to get its embedding in the current layer, i.e., residual connection. 

\subsection{Heterogeneous Message Passing}
To handle the graph heterogeneity caused by different types of nodes and connections/edges, we adopt a heterogeneous message-passing mechanism where node type-dependent modules are used to map feature distributions of different types of nodes into the same latent. Then edge type-dependent modules are used to handle the edge heterogeneity. Specifically, the messages of passed from neighbors $\{v_i\}_{i=1}^N$ to the target node $v$ in the $l$-th layer is formulated as
\begin{equation} \label{eq:message}
\begin{aligned} 
\bm{h}^{\text{msg}}_{v, v_i} &= \text{FFN}^V_{\phi(v_i)}(\bm{h}^{t_i, l-1}_{v_i})W^{\text{msg}}_{\varphi(e_i)}, i \in [1,N] , 
\end{aligned}
\end{equation}
where $\bm{h}^{\text{msg}}_{v, v_i}$ represents the transformed message passed from $v_i$ to $v$, 
$N$ is the number of neighbours, $\text{FFN}^V_{\phi(\cdot)}$ and $W^{\text{msg}}_{\varphi(\cdot)}$ are the node type- and edge type-independent transformation modules. $\bm{h}^{t_i, l-1}_{v_i})$ denotes the representation of $v_i$ from the $l-1$-th layer ($l \ge 1$) at time $t_i$. Moreover, $e_{i}$ denotes the edge between node $v$ and $v_i$, $\phi(v)$ and $\varphi(e)$ denote the types of node $v$ and edge $e$, respectively. Note that we use bold font to denote vector variables. 

\subsection{Local Aggregation on Temporal HIN}
As mentioned in Sec.~\ref{model_overview}, we incorporate three key factors in the aggregation process of a CTRL layer which are handled by the following sub-modules. 

% \subsection{Heterogeneous Graph Attention}
\textbf{Heterogeneous Attention}
The attention scores, calculated as the normalized scaled-dot product values between the representations of the target node and its neighbours, measure the semantic correlation between nodes and are widely used in the aggregation process of GNN, e.g., GAT and SAGE. 
% Generally, high attention values are given to semantically important nodes. 
% Semantically important nodes will be assigned with high attention values. 
However, in HINs the target node and its neighbours may have different feature distributions. 
% due to heterogeneous node types. 
% Moreover, there are also heterogeneous relations/edges between nodes.
Moreover, there may exist different types of edges between a pair of node types. 
Therefore, edge information between two nodes should be considered when calculating their semantic correlation.  
% Therefore, we need to further incorporate the edge information between two nodes when calculating their semantic correlation.  

Specifically, for an edge, we first map the representations of heterogeneous nodes into the same latent space via node type-dependent mapping functions to handle the node heterogeneity. 
Then, the mapped representations of the source nodes are further transformed through an edge type-dependent matrix to further incorporate the edge heterogeneity. 
Finally, the attention score is computed by the scale-dot product between the transformed representations of the source and target nodes. 
Therefore, given a target node $v$ at timestamp $t$ with its neighbor set $\{(v_i)\}^N_{i=1}$, the attention score is calculated by
% their hidden representations from the $(l-1)$-th layer, i.e., $h^(l-1)_{v_t}$ and $h^(l-1)_{v_n}$,
\begin{equation} \label{eq:attn-content}
\begin{aligned} 
Q &= \text{FFN}^Q_{\phi(v)}(\bm{h}^{t, l-1}_{v}),  \\
K &= \{\text{FFN}^K_{\phi(v_i)}(\bm{h}^{t_i, l-1}_{v_i}) W^{Key}_{\varphi(e_{i})}\}_{i=1}^N, \\
\text{Attn}^{t,v} &= \{\text{attn}^{t,v}_{v_i}\}_{i=1} ^N = \text{softmax}(\frac{QK^T}{\sqrt{d}}),
\end{aligned}
\end{equation}
where $d$ is the dimension of the hidden representations, $\text{FFN}^Q_{\phi(\cdot)}$ and $\text{FFN}^K_{\phi(\cdot)}$ are the node type-dependent feedforward network for the \emph{Query} and \emph{Key} terms, respectively. $W^{Key}_{\varphi(\cdot)}$ is the edge type-dependent matrix which further transfers the representations of neighbour nodes based on their connection types between the target node. 

% \subsection{Edge-based Hawkes Process}
\textbf{Edge-based Hawkes Process.}
% Intuitively, the impact of historical node decay over time.  Previous methods often leverage the Hawkes process to model this temporal impact between nodes. However, in a temporal HIN, different types of historical nodes may have different dynamics, i.e. different decay rates. 
Intuitively, the impact of historical nodes decays over time.  Previous methods often leverage the Hawkes process to model this temporal impact between nodes. However, in a temporal HIN, different types of historical nodes may have different dynamics, i.e. different decay rates.  
Furthermore, the temporal decay of impact may vary even for the same type of node. 
For example, in a social network, the impacts of different celebrities may decay differently for different followers depending on their connection(edge) types, e.g., mutual follow or unilateral follow. 

% that is closer in time should have a greater impact. 
To this end, we propose an edge-based Hawkes process to measure the influence of neighbour nodes according to their connection information with the target node. 
To be specific, all the available information of an edge, including types and representations of the target and source nodes and also the type and feature (if available) of the edge, are taken into account to generate the edge-specific decay rate for the corresponding edge-based Hawkes process.  
% we take node representations and graph heterogeneities into account to generate an edge-specific decay rate.  
% To better model the dynamic impact between heterogeneous nodes. We further integrate the edge-based Hawkes process into the GNN layer. 
% The edge-based Hawkes process learns an edge-specific decay rate for every interaction between nodes based on their temporal hidden representations. 
Formally, the influence intensities for a set of historical neighbour nodes $\{v_i\}_{i=1}^N$ of a target node $v$ at time $t$ are calculated by
\begin{equation} \label{eq:hawkes}
\begin{aligned}
\bm{h}^q_v &= \text{FFN}^Q_{\phi(v)}(\bm{h}^{t, l-1}_{v}), \\ 
\bm{\bar{h}}^k_{v_i} &= \text{FFN}^K_{\phi(v_i)}(\bm{h}^{t_i, l-1}_{v_i}) W^{Key}_{\varphi(e_{i})}, \\ %i \in [1,N] \\ 
\delta_{e_i} &= \sigma(\text{MLP}(\bm{h}^q_{v} || \bm{\bar{h}}^k_{v_i})), \\ %i \in [1, N]\\
\lambda^{v,t} &= \{\lambda^{v,t}_{v_i}\}_{i=1}^N = \text{softmax}(\{ \kappa(t-t_i, \delta_{e_i}) \}_{i=1}^N) ,
\end{aligned}
\end{equation}
where \red{$e_{i}$ represents the edge between the target node $v$ and neighbor $v_i$, 
% $h^{t_i, l-1}_{v_i}$ denotes the hidden representation of node $v_i$ at time $t_i (< t)$ from the $(l-1)$-th layer, 
$t_i (< t)$ denotes the time when the edge $e_i$ between node $v$ and $v_i$ is established, 
and $||$ denotes the concatenation operation between two vectors.  
Moreover, introduced in~\eqref{eq:attn-content}, $\text{FFN}^Q_{\phi(\cdot)}$ and $\text{FFN}^K_{\phi(\cdot)}$ are node type-dependent modules to handle the node heterogeneity and $W^{Key}_{\varphi(\cdot)}$ is edge type-dependent matrix to deal with the heterogeneous connections. 
% Furthermore, the output of the $\text{MLP}$ module, i.e., $\delta_{e_i}$, denotes the decay rate of the impact of neighbor $v_i$ which is connected with the target node via edge $e_i$. 
Furthermore, the $\text{MLP}$ module takes the transformed latent representations of the target and source nodes, i.e., $\bm{h}^q_v$ and $\bm{\bar{h}}^k_{v_i}$, as input and outputs the decay rate, $\delta_{e_i}$, of the Hawkes process that models the dynamic impact of neighbour $v_i$ to $v$. 
Introduced in ~\eqref{eq:hawkes_org}, $\kappa(\Delta t, \delta) = exp(-\delta \Delta t)$ calculates the dynamic influence intensity. 
% the output of the $\text{MLP}$ module, i.e., $\delta_{e_i}$, denotes 
Note that, $\sigma(\cdot)$ is the ReLU~\cite{agarap2018deep} activation function.} 
Finally, the relative temporal influence of all historical neighbour nodes is captured with the softmax function.
% on the 

% \subsection{Dynamic Centrality} 
\textbf{Dynamic Centrality} 
As shown in Figure~\ref{fig:sys}, we first get the dynamic degree centrality of a node at a given timestamp, e.g., the dynamic degree of node $P_1$ at $t_3$ is denoted as $\text{Degree}(P_1, t_3)$. 
% influence in the aggregation. 
% Then, we also use the dynamic node degree to weigh the importance when aggregating the representations of neighbours.
The dynamic degree measures the node importance at the current time which can is incorporated into the aggregation procedure of the CTRL layer. 
However, the degrees of different types of nodes are often not comparable. For example, in a citation network, the dynamic degree of a venue (or conference) is much larger than the degree of an author. 
Therefore, we adopt trainable node-type dependent variables, i.e., $\beta_{\phi(\cdot)}$, to scale the degrees of different types of nodes before feeding into the softmax function. To be specific, given a target node $v$ and its neighbors $\{v_i\}_{i =1}^N$ at time $t$, we have
\begin{equation} \label{eq:node_weight}
\begin{aligned}
 \omega^{v,t} = \{\omega^{v,t}_{v_i} \}_{i=1}^N = \text{softmax}(\{ \beta_{\phi(v_i)} * D^t_{v_i} \}_{i=1}^N),
\end{aligned}
\end{equation}  
where $D^t_{v_i}$ denotes the dynamic degree centrality of node $v_i$ at $t$.
% and $\beta_{\phi(\cdot)}$ is a node type-dependent trainable scalar that is used to scale the dynamic node degree 

\red{Finally, the hidden representation of node $v$ at time $t$ in the $l$-the CTRL layer is formulated as}
\begin{equation} \label{eq:gnn}
\begin{aligned}
% \resizebox{0.95\columnwidth}{!}{
\bm{h}^{\text{msg}}_v &= \sum_{i=1}^N (\alpha_1 \text{attn}_{v_i}^{v, t} + \alpha_2 \lambda_{v_i}^{v, t} + \alpha_3 \omega_{v_i}^{v, t}) \bm{h}^{\text{msg}}_{v, v_i} , \\ 
\bm{h}^{t, l}_{v} &= \bm{h}^{t, l-1}_v + \text{Adapt}_{\phi(v)}( \bm{h}^{\text{msg}}_v) ,
% }
\end{aligned}
\end{equation}  
% where $\bm{h}^{t, l-1}_v, l \ge 1$ is the hidden representation of node $v$ at time $t$. 
where $\bm{h}^{\text{msg}}_{v,v_i}$, defined in~\eqref{eq:message}, is the transformed message passed from $v_i$ to $v$, 
$\bm{h}^{\text{msg}}_v$ is the aggregated message received by node $v$ from all its neighbor nodes. 
$\alpha_1 + \alpha_2 + \alpha_3 = 1$ are trainable weights for the semantic importance, temporal influence, and the dynamic degree centrality of each neighbour. $\text{Adapt}_{\phi(v)}$ is the node type-dependent adapt module, which consists of a single linear layer, that maps the received message vector, $\bm{h}^{\text{msg}}_v$, to the latent feature space of the target node $v$. 
Note that we only consider undirected graphs here, however, the degree of centrality can be easily extended to directed graphs.

% \red{Then, we explore the dynamic degree centrality of heterogeneous nodes from two perspectives to better understand a temporal HIN.  First, we treat the dynamic degree of a node as its dynamic feature. Thus, we develop a degree encoding module which assigns each node a real-valued vector to its dynamic degree. The degree embedding is further added to the node feature as the input of the first GNN layer. That is,}
Furthermore, the dynamic degree is also an important dynamic feature of nodes.  
Thus, we develop a degree encoding module which assigns real-valued vectors to the dynamic degree of each node. 
The degree embedding is further added to the node feature as the input of the first CTRL layer. That is,
\begin{equation} \label{eq:node_embed}
\bm{h}^{t, 0}_{v_i} = \bm{x}_{v_i} + \bm{z}_{v_i,t} ,
\end{equation}  
where $\bm{x}_{v_i}$ is the node feature of node $v_i$, $\bm{z}_{v_i, t}$ is the embedding of the dynamic node degree of node $v_i$ at time $t$. For a directed graph, $\bm{z}_{v_i, t}$ can be divided into $\bm{z}^{+}_{v_i, t}$ and $\bm{z}^{-}_{v_i, t}$ which stand for the embeddings of the dynamic indegree and outdegree.

% Furthermore, the original feature of node $v$, i.e., $\bm{h}^{t, 0}_v$, is calculated in~\eqref{eq:node_embed}.

\subsection{Event-based Training}
% on how to get node embeddings that effectively capture the high-order local topological dynamics of temporal HINs
Previous temporal network methods are often trained by modelling the temporal link formation process which ignores the evolution of high-order network structures. 
% Therefore, we propose to train the CTRL by modelling the occurrence of the basic event in a given temporal HIN. 
Therefore, we propose to train the CTRL by modelling the evolution of the events in a given temporal HIN. 

Different networks have different basic events, for example, the event of a citation network should be the \emph{publication of a paper} which is a subgraph including the authors, venue, and the cited papers. 
Therefore, we first identify the basic event for a given temporal HIN and model the evolution of the predefined basic even from two perspectives.  

First, we predict the probability of an event, as a whole, occurring. 
To be specific, for an event, which is subgraph, $\mathcal{G}^t_s = \{\mathcal{V}^t_s, \mathcal{E}^t_s\}$ at time $t$.
We get the event embedding by taking averages of the node representations in the event, that is, 
% We get the embeddings of both subgraphs by taking averages of their node representations, that is, 
\begin{equation} \label{eq:graph_embed}
\begin{aligned}
\bm{h}_{g^t_s} = \frac{1}{|\mathcal{V}^t_s|}\sum_{v \in \mathcal{V}^t_s} \bm{h}^{t, L}_{v} ,
\end{aligned}
\end{equation} 
where $\bm{h}^{t, L}_{v}$ is the latent node representation from the last CTRL layer. 
Then, we build an MLP module, $\text{MLP}_{\text{event}}$, which takes the event representation as input and predicts the probability of an event occurring. 
% the to predict the probability of the occurrence of a subgraph. 
% that is, $p^t_{g_s} = \text{MLP}_g(\bm{h}^t_{g_s})$. 
\begin{equation} \label{eq:graph_prob}
\begin{aligned}
p_{g^t_s} = \text{MLP}_{\text{event}}(\bm{h}_{g^t_s}) .
\end{aligned}
\end{equation} 

Secondly, to preserve the topological information within each event, we further predict the occurrence probability of each edge in an event. 
Specifically, we further build another MLP module, $\text{MLP}_{edge}$, that takes the representations of the source and target node of an edge as input and outputs the probability of the edge occurs. Formally, for an given $e^t_j = \{v_i, v_j, t\} \in \mathcal{E}^t_s$, the edge probability is calculated as: 
\begin{equation} \label{eq:edge_prob}
p_{e^t_k} = \text{MLP}_{\text{edge}}(\bm{h}^{t,L}_{v_i} || \bm{h}^{t,L}_{v_j}).
\end{equation}

% the structure within an event should be preserved and taken into account in the training process. 
% Therefore, we further build another MLP layer that takes the representations of the source and target node of an edge as input and outputs the probability of the edge. 
% Specifically, for each edge $e \in \mathcal{E}^t_s$ with node $v_i$ and $v_j$, we first sample a negative edge $e' = \{v_i, v'_j\}$ and calculate the edge-based loss as 

To explore the evolution of the whole network, for each event $\mathcal{G}_s^t $, we apply negative sampling on the temporal HIN to create a negative event $\mathcal{G}^t_n = \{\mathcal{V}^t_n, \mathcal{E}^t_s\}$ by randomly sample a set of negative nodes $\mathcal{V}^t_n$. 
Note that, there are still overlapping between the node sets of the $\mathcal{G}_s^t $ and $\mathcal{G}_n^t$, i.e., $\mathcal{V}^t_s \cap \mathcal{V}^t_n \ne \emptyset $. 
Then, the event occurrence loss of $\mathcal{G}^t_s$ is calculated as:  
\begin{equation} \label{eq:loss_graph}
\begin{aligned}
\mathcal{L}^{g^t_s}_{\text{occur}} = -log(p_{g^t_s}) - log(1 - p_{g^t_n}),
\end{aligned}
\end{equation} 
where $p_{g^t_n}$ is the occurrence probability of the negative event $\mathcal{G}_n^t$ defined in~\eqref{eq:graph_prob}. 
% Furthermore, the structure within an event should be preserved and taken into account in the training process. 
% Therefore, we further build another MLP layer that takes the representations of the source and target node of an edge as input and outputs the probability of the edge. 
% Specifically, for each edge $e \in \mathcal{E}^t_s$ with node $v_i$ and $v_j$, we first sample a negative edge $e' = \{v_i, v'_j\}$ and calculate the edge-based loss as 
Furthermore, for each edge $e^t_k = \{v_i, v_j, t \} \in \mathcal{E}^t_s$, we create a negative edge $\hat{e}^t_k = \{v_i, v'_j, t \}$ by randomly sample a negative target node $v'_j$. 
Then the structural loss of event $\mathcal{G}_s^t$ is computed by: 
\begin{equation} \label{eq:loss_edge}
\mathcal{L}^{g^t_s}_{\text{topo}} = - \frac{1}{|\mathcal{E}^t_s|}\sum_{e^t_k \in \mathcal{E}^t_s}(log(p_{e^t_k}) + log(1 - p_{\hat{e}^t_k})),
\end{equation}
where $p_{\hat{e}^t_k}$ is the probability of negative edge $ \hat{e}^t_k$ defined in~\eqref{eq:edge_prob} and $\mathcal{E}^t_s$ is the edge set of event $\mathcal{G}^t_s$. 

Finally, we optimize model parameters $\theta$ through the following objective function 
\begin{equation} \label{eq:loss_final}
\begin{aligned}
\argmin_{\theta} \sum_{\mathcal{G}^t_s \in \mathcal{G}} \mathcal{L}^{g^t_s}_{\text{occur}} + \mathcal{L}^{g^t_s}_{\text{topo}}, 
% l = l_g + \frac{1}{|\mathcal{E}^t_s|}\sum_{e \in \mathcal{E}^t_s } l_e
\end{aligned}
\end{equation}
where $\mathcal{G}$ is a temporal HIN. 
% We train the model with a mini-batch of the predefined basic events 
We optimize the objective with gradient descent on a batch of predefined events. 

\begin{table}[tbp]
% \small
\begin{center}
\resizebox{\columnwidth}{!}{
\begin{tabular}{ccccc}
\toprule
Datasets  & \#N/E-Types & \#Nodes & \#Edges  & Time Range\\
\hline
ACM  & 3; 3 & 87,926  & 204,436  & 2000-2016 \\
DBLP & 4; 4 & 147,138 & 612,673 & 2010-2020  \\
IMDB & 2; 6 & 96,175  & 217,358 & 2000-2020  \\
\bottomrule
\end{tabular}
}
\caption{Data statistics. \#N/E-Types denotes the number of node and edge types.}  
\label{table:dataset_info}
\end{center}
\end{table}

\section{Experiments}
\label{sec:exp}

We conduct extensive experiments and compare our approach with state-of-the-art methods to demonstrate the effectiveness of the proposed model.  

\begin{table*}[tbp]
% \small
\begin{center}
\resizebox{\textwidth}{!}{
\begin{tabular}{c|ccc|ccc|ccc}
\toprule
\multirow{2}{*}{Methods}     & \multicolumn{3}{c|}{ACM}  & \multicolumn{3}{c|}{DBLP} & \multicolumn{3}{c}{IMDB}    \\
\cline{2-10}                          
                             &  Accuracy(\%) & AP(\%) & F1(\%) &  Accuracy(\%) & AP(\%) & F1(\%) &  Accuracy(\%) & AP(\%) & F1(\%) \\
\midrule
GCN     & 58.79±0.3 & 67.29±1.02 & 65.72±0.16 & 66.59±0.18 & 72.62±0.24 & 67.81±0.16 & 59.26±0.22 & 62.38±0.41 & 59.62±0.73 \\
SAGE    & 60.7±1.21 & 63.9±1.18 & 66.21±0.95 & 67.47±0.16 & 73.57±0.14 & 73.54±0.13 & 61.35±0.56 & 67.59±1 & 61.86±1.86 \\
GAT     & 61.35±0.74 & 63.96±1.42 & 63.3±1.08 & 68.84±0.47 & 75.91±0.65 & 67.69±0.65 & 61.28±0.12 & 65.6±0.4 & 62.05±1.48 \\
RGCN    & 63.37±0.35 & 65.35±0.42 & 63.94±0.62 & 68.8±0.32 & 73.98±0.43 & 68.31±0.48 & 67.33±0.46 & 71.64±0.7 & 67.12±1.21  \\
\hline
TGAT    & 72.77±1.47 & 79.95±1.13 & 71.43±1.5 & 75.09±2.11 & 82.81±2.35 & 72.93±2.95 & 82.33±0.3 & 88.63±0.45 & 83.14±0.25 \\
HGT     & 68.01±0.36 & 71.78±1.24 & 68.46±0.43 & 69.65±0.42 & 77.49±0.19 & 67.84±0.93 & 67.35±0.23 & 72.77±0.23 & 68.48±0.27 \\
TGSRec  & 75.02±1.29 & 73.93±3.95 & 78.08±1.07 & 79.76±0.28 & 86.5±0.08 & 80.33±0.63 & 75±0.73 & 83.47±1.67 & 77.66±0.91    \\
CAW     & 73.87±0.09 & 84.64±0.14 & 76.64±0.09 & 74.95±0.16 & 86.68±0.11 & 70.21±0.56 & 74.66±0.03 & 85.24±0.02 & 73.45±0.17 \\
TREND   & 55.72±1.53 & 53.15±0.94 & 59.54±3.16 & 54.69±2.44 & 52.49±8.84 & 64.11±9.1 & 52±0.24 & 51.04±2.23 & 55.09±2.43 \\
\midrule
CTRL & \textbf{84.51±0.12} & \textbf{91.86±0.57} & \textbf{85.52±0.06} & \textbf{86.46±0.08} & \textbf{94.15±0.03} & \textbf{86.22±0.08}  & \textbf{90.91±0.25} & \textbf{95.69±0.39} & \textbf{91.19±0.12} \\ 
Imp.(\%) & 12.65\% & 8.53\% & 9.53\% & 8.4\% & 8.62\% & 7.33\% & 10.42\% & 7.97\% & 9.68\% \\
% 8.4\%	8.62\%	7.33\%
\bottomrule
\end{tabular}
}
\caption{Results of the inductive temporal link prediction task on the three public datasets. Imp.\% indicates the relative performance improvement of our methods compared to the best results given by all the baselines. All improvements are significant with a t-test p-value less than 0.05.}
\label{table:main_result}
\end{center}
\end{table*}

\subsection{Experimental Setup}
\textbf{Datasets.} We evaluate the proposed approach and all the baseline methods on three real-world datasets, i.e., ACM\footnote{https://www.aminer.cn/citation\#b541}, DBLP, and IMDB\footnote{https://www.imdb.com/interfaces/}. 
The ACM and DBLP are two paper citation networks where the ACM dataset contains three types of nodes (``paper'', ``author'', and ``venue'') and the DBLP dataset involves four types of nodes (``paper'', ``author'', ``venue'', and ``field''). 
The IMDB dataset is a network with two types of nodes (``movie'' and ``people'') and six types of edges (``editor'', ``actress'', ``actor'', ``director'', ``writer'', and ``producer'').  
\footnote{Dataset details can be found in supplementary}The statistics of these datasets are shown in Table~\ref{table:dataset_info}. 

% \red{node degree}
% For the citation networks, namely, the ACM and DBLP datasets, we use Doc2vec~\cite{le2014distributed} model to convert the title and abstract of each paper into a real-valued vector with a dimension size of 128 which is treated as the raw feature of a paper. 
For citation networks, i.e., ACM and DBLP datasets, we use Doc2vec~\cite{le2014distributed} model to convert the title and abstract of each paper into a real-valued vector with a dimension size of 128, which is treated as the raw features of the paper node.
Moreover, in the DBLP dataset, we leverage the word2vec method to convert each ``field'' into a vector of length 128. 
Then, we use all-zero vectors as the raw features of other node types, i.e., ``author'' and ``venue'' nodes.
For the IMDB dataset, we encode the raw features of a movie, including its title, genre, and release region, into a vector of size 246 and also use all-zero vectors as the raw features of  ``people'' nodes.   
% For the IMDB dataset, we encode the raw features of a movie, including its title, genre, and release region, into a vector of size 246 and also use all-zero vectors for ``people'' nodes.   

% In addition, the basic event for a citation network is defined as \emph{the publication of a paper} which includes the authors of the paper, the venue where the paper is published, the cited papers, and the field of study (available for the DBLP dataset). For the IMDB dataset, the basic event is defined as \emph{the release of a movie} consists the movie node and corresponding crews including the director, producer, actor, editor etc. 
In addition, the basic event for a citation network is defined as \emph{the publication of a paper} which includes the authors of the paper, the venue where the paper is published, the cited papers, and the field of study (available for the DBLP dataset). For the IMDB dataset, the basic event is defined as \emph{the release of a movie} consisting of the movie node and its corresponding crews including director, producer, actor, editor, etc. 

\begin{table}[tbp]
% \small
\begin{center}
\resizebox{\columnwidth}{!}{
\begin{tabular}{cccc}
\toprule
Models  & HIN    & Temporal &  Centrality  \\ % & Time Range\\
\hline
GCN~\cite{kipf2017semi}                & \ding{56} & Sample   & \ding{56} \\ 
SAGE~\shortcite{hamilton2017inductive} & \ding{56} & Sample   & \ding{56} \\ 
GAT~\cite{velickovic2018graph}         & \ding{56} & Sample   & \ding{56} \\ 
RGCN~\cite{schlichtkrull2018modeling}  & \ding{56} & Sample   & \ding{56} \\ 
\hline
TGAT~\cite{tgat_iclr20}                & \ding{56} & Encode    & \ding{56} \\ 
HGT~\cite{hgt}                         & \ding{52} & Encode    & \ding{56} \\
TGSRec~\cite{fan2021continuous}        & \ding{52} & Encode    & \ding{56} \\
CAW~\cite{wang2021inductive}           & \ding{56} & Encode    & \ding{52} \\ 
TREND~\cite{wen2022trend}              & \ding{56} & Hawkes    & \ding{52} \\
\bottomrule
\end{tabular}
}
\caption{Details of baseline methods. The ``Temporal'' column indicates how each baseline utilizes time information where ``Sample'' for temporal sampling, ``Embed'' for time encoding, and ``Hawkes'' for Hawkes process.}  
\label{table:baselines}
\end{center}
\end{table}

\noindent \textbf{Experimental setup.}
% We split a temporal HIN into three parts for training, validation, and testing, according to the timestamps to train all the methods and evaluate them with an inductive temporal link prediction task. 
% The data split is performed according to the occurrence time of the predefined basic event in each network. That means all the events in the validation and testing sets are not involved in the training process and all the edges in the validation and testing sets are new edges where at least one node of an edge is not involved in model training. 
% We evaluate all methods with four metrics including Accuracy, Averaged Precision (AP), F1 score, and AUC. 
% We repeat all experiments at least three times and give out the average and standard deviation of all metrics.
We split a temporal HIN into three parts for training, validation, and testing, according to the timestamps to make sure that all methods are trained and evaluated under the same conditions. 
Specifically, we split up a temporal network according to the occurrence time of the predefined basic event. 
Therefore, all the events in the validation and testing sets are not involved in the training process and all the edges in the validation and testing sets are new edges where at least one node of an edge is not involved in model training.  So that we can evaluate all methods with the inductive link prediction task. 
We evaluate all methods with four metrics including Accuracy, Averaged Precision (AP), F1 score, and AUC.
We repeat all experiments at least three times and give out the average and standard deviation of all metrics.

\noindent \textbf{Baselines.}
We compare the proposed method with a range of state-of-the-art network embedding methods. 
%, including heterogeneous graph embedding and temporal graph embedding models. 
Specifically, we have compared with the baseline methods in Table~\ref{table:baselines}. 
For static methods, i.e., GAT, SAGE, GCN, and RGCN, we use temporal neighbourhood sampling to adapt to the temporal graph embedding setting. 

\subsection{Implementation Details}
We evaluate the proposed method and baselines on a server with AMD EPYC 7K62 48-Core Processor and NVIDIA Tesla P40 GPUs. The experimental environment is Linux 3.10 with CUDA 10.1. 
To have fair comparisons, the dimension of node embedding is set to 128 for all methods. We use the Adam optimizer with a learning rate of 0.001 and adopt a batch size of 1024. 
Moreover, we set the number of neighbours $N=10$ and randomly select one negative sample for a positive edge in both training and evaluation. 
We apply two GNN layers for both the CTRL model and all baseline methods. 
The number of heads of the scaled-dot product module is set to 2. 
In the training stage of the CTRL, 
we sampled (with repeat) 5 authors, 10 cited papers, and 10 fields of study for each paper to form the ``publication of a paper'' event in the ACM (without the ``field'' node) and DBLP datasets. Similarly, 10 ``people'' nodes with different roles, e.g, director, actor, etc., are sampled for each movie to form the ``movie release'' event in the IMDB dataset. 
Furthermore, we use the default settings for other parameters of the baselines. 

\begin{figure}[tbp]
\centering
\includegraphics[width=0.75\columnwidth]{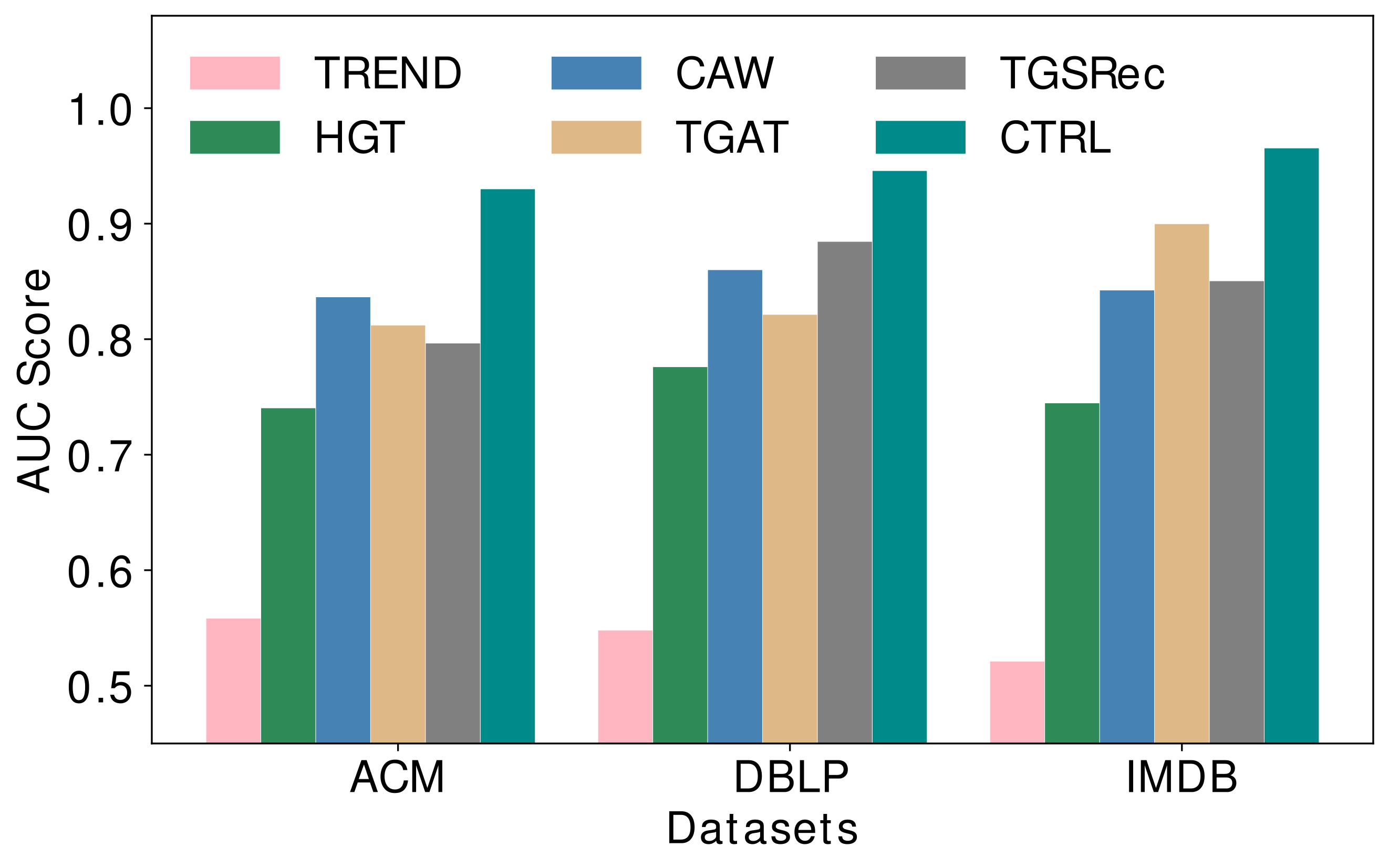} % Reduce the figure size so that it is slightly narrower than the column. Don't use precise values for figure width. This setup will avoid overfull boxes.
\caption{\red{AUC of the proposed method and temporal network embedding baselines on the three datasets.}}
\label{fig:baseline_auc}
\end{figure}

\begin{table*}[tbp]
\begin{center}
\resizebox{\textwidth}{!}{
\begin{tabular}{c|ccc|ccc|ccc}
\toprule
\multirow{2}{*}{Variants}     & \multicolumn{3}{c|}{ACM}  & \multicolumn{3}{c|}{DBLP} & \multicolumn{3}{c}{IMDB}    \\
\cline{2-10}                          
                             &  Accuracy(\%) & AP(\%) & F1(\%) &  Accuracy(\%) & AP(\%) & F1(\%) &  Accuracy(\%) & AP(\%) & F1(\%) \\
\midrule
CTRL            & \textbf{84.51±0.12} & \textbf{91.86±0.57} & \textbf{85.52±0.06} & \textbf{86.46±0.08} & \textbf{94.15±0.03} &  \textbf{86.22±0.08}  & \textbf{90.91±0.25} & \textbf{95.69±0.39} & \textbf{91.19±0.12} \\ 
-Event\_loss    & 83.56±0.88 & 90.91±0.76 & 84.79±0.7 & 86.07±0.31 & 93.74±0.24 & 85.87±0.32 & 89.3±0.14 & 93.91±0.05 & 89.81±0.08 \\
-Centrality     & 82.47±0.29 & 89.64±0.25 & 83.95±0.26 & 84.63±0.13 & 91.81±0.16 & 84.79±0.32 & 88.4±0.02 & 94.05±0.26 & 88.54±0.04 \\
Hawkes-$\delta$ & 81.01±0.37 & 88.97±0.16 & 81.56±0.37 & 83.49±0.27 & 91.41±0.21 & 83.37±0.31 & 88.19±0.18 & 94.03±0.1 & 88.37±0.14 \\
\bottomrule
\end{tabular}
}
\caption{Results of ablation studies on the three public datasets.}
\label{table:ablation_result}
\end{center}
\end{table*}

\subsection{Main Results}
Table~\ref{table:main_result} summarizes the experimental results of all baselines and the proposed method on the three real-world temporal HINs. 
The proposed model significantly outperforms all other baselines with a relative performance improvement on accuracy, average precision (AP), and F1 scores of 12.65\%, 8.53\%, and 9.53\% on the ACM dataset. Similarly, on the DBLP and IMDB datasets, we achieved a relative improvement of 8.4\%, 8.62\%, 7.33\% and 10.42\%, 7.97\%, 9.68\%. In addition, Figure~\ref{fig:baseline_auc} shows the AUC scores of our model and other temporal graph embedding baseline. We observe that the proposed model achieved AUC scores over 0.9 in all datasets, which significantly outperforms the baselines. 

Generally, the temporal methods are able to outperform all the static methods except for the TREND model. In the training stage, TREND tries to predict the dynamic node degree which is problematic in some networks, e.g., in citation networks, the dynamic degree of the ``venue'' node is on the order of thousands, while the dynamic degree of ``author'' node is less than 10 for most authors. Moreover, in the movie network, the dynamic degree of a movie is also 0. These may cause the network to focus more on the node degree and fail to preserve the network structure. 
In our work, the dynamic degree is encoded into node features and is scaled with trainable type-dependent variables. Therefore, the proposed model can capture dynamic centrality without such a problem.

\begin{figure}[tbp]
\centering
\includegraphics[width=0.75\columnwidth]{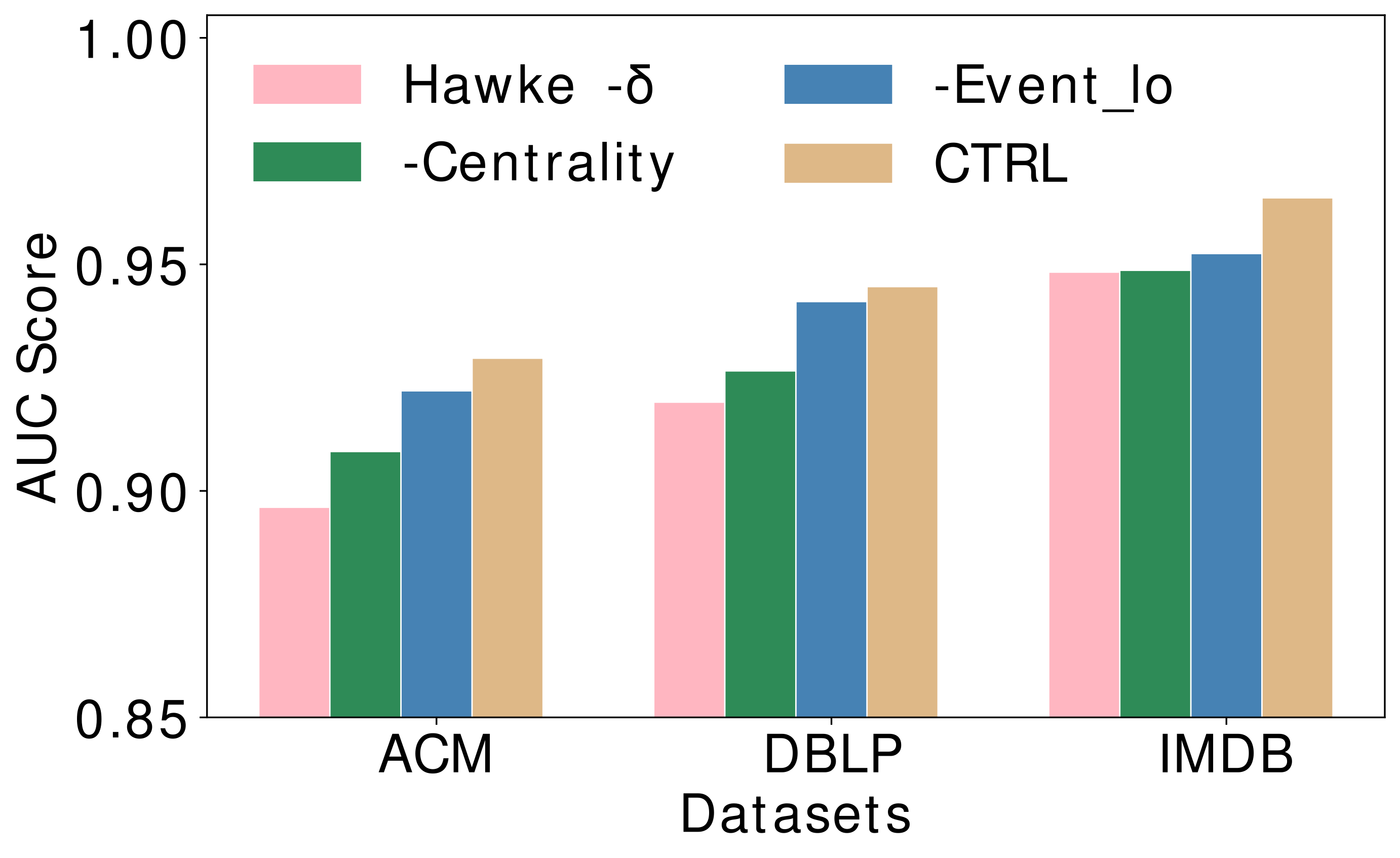} % Reduce the figure size so that it is slightly narrower than the column. Don't use precise values for figure width. This setup will avoid overfull boxes.
\caption{AUC scores of ablation studies.
% on the three datasets.
}
\label{fig:ablation_auc}
\end{figure}

\subsection{Ablation Study}
% Variants of the proposed method are evaluated on the collected datasets for ablation studies to show the impact of each proposed sub-module We incrementally accommodate different modules until we incorporate all the proposed sub-modules and features. Specifically, the following models are evaluated:
We further conduct ablation studies where variants of the proposed method are evaluated to demonstrate the effectiveness of each proposed module and feature. We gradually remove different sub-modules or features from the CTRL model and show the impact of each module. 
Specifically, apart from the proposed method, i.e., \textbf{CTRL}, the following variants are evaluated:
\begin{itemize}[leftmargin=*]
    % \item \textbf{CTRL.} The proposed method. 
    \item \textbf{-Event\_loss.} We remove the event loss and train the model with edge loss, the same as previous works, TGAT, CAW, etc. 
    \item \textbf{-Centrality.} We further remove the dynamic centrality sub-module. 
    \item \textbf{Hawkes-$\delta$.} We further replace the edge-based Hawkes process with the original Hawkes process where a single trainable decay rate, i.e., $\delta$, is used like TREND.
    % ~\cite{wen2022trend}.
\end{itemize}

Table~\ref{table:ablation_result} summarizes the results of the accuracy, average precision, and F1 score of ablation studies. The AUC score is plotted in Figure~\ref{fig:ablation_auc}. We observe performance degradation each time a sub-module or feature is removed from the previous model. Specifically, the comparison between the \textbf{CTRL} and the \textbf{-Event\_loss} model demonstrates the effectiveness of event-based training for capturing the high-order evolution of temporal HINs. 
Moreover, the better performance achieved by the \textbf{-Event\_loss} compared to the \textbf{-Centrality} indicates the necessity of integrating the dynamic centrality in the temporal graph. Finally, comparison results between the \textbf{-Centrality} and the \textbf{Hawkes-$\delta$} model show the advancement of the proposed edge-based Hawkes process over the original Hawkes process on the temporal HINs. 
% further analysis. 

\section{Conclusion}

In this paper, we propose the CTRL model for continuous-time representation learning on temporal HINs. In the message passing stage of a CTRL layer, node type- and edge-type-dependent parameters are used to handle the graph heterogeneity. Moreover, in the aggregation steps, we consider the semantic correlation, temporal influence, and dynamic node centrality to determine the importance of neighbour nodes. Finally, we train CTRL with a future event prediction task to capture the evolution of high-order network structure. Extensive experiments on three real-world datasets demonstrate the superiority of CTRL and the effectiveness of the model design.

%!TEX root = ../main.tex

\section{Appendix}

\subsection{Dataset Details}

\begin{table}[h]
\begin{center}
\resizebox{\columnwidth}{!}{
\begin{tabular}{ccccc}
\toprule
Datasets  & Node Types & \#Nodes & Edges Types & \#Edges \\
\hline
\multirow{3}{*}{ACM} & Paper  & 37181 & Author-Paper & 114649 \\ 
                     & Author & 50731 & Paper-Paper & 52610   \\
                     & Venue  & 14    & Paper-Venue & 37181    \\
\hline                     
\multirow{4}{*}{DBLP} & Paper  & 53682 & Author-Paper & 194389 \\ 
                     & Author & 91051 & Paper-Paper & 136918   \\
                     & Venue  & 14    & Paper-Venue & 53682    \\
                     & Field  & 2391  & Paper-Field & 227684    \\
\hline                     
\multirow{3}{*}{IMDB} & Movie  & 44555 & Editor, Actress & 11209, 41940   \\
                      & People & 51620 & Actor, Director & 75115, 31318   \\
                      &        &        & Writer, Producer & 25121, 32655   \\
\bottomrule
\end{tabular}
}
\caption{Data statistics.}  
\label{table:dataset_all}
\end{center}
\end{table}

\begin{figure}[h]
    \centering
    % \begin{subfigure}
    \subfigure[Papers published in each venue.]{\includegraphics[width=3in]{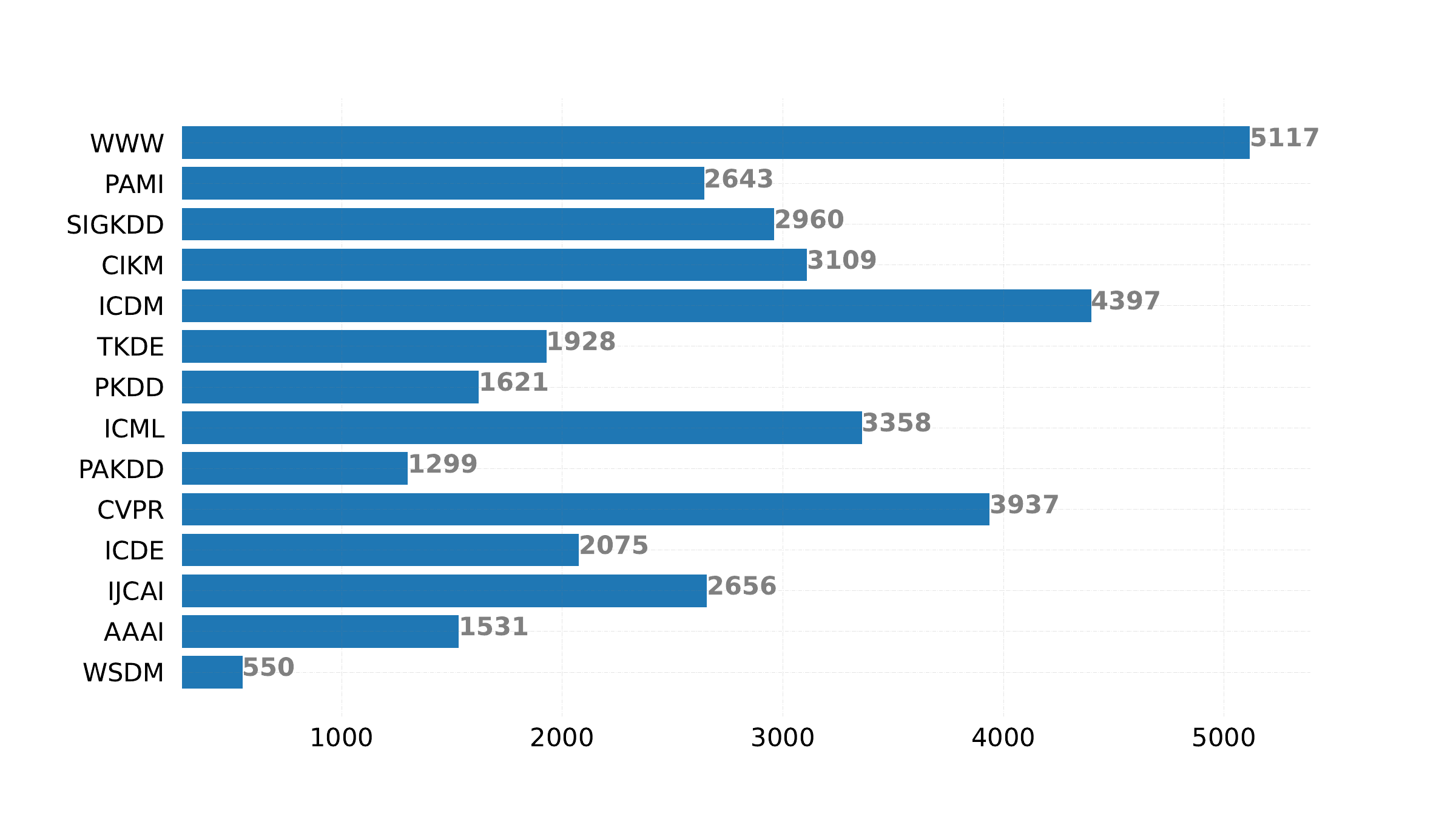}}
        % \caption{Papers published in each venue.}
    % \end{subfigure}
    
    \bigskip
    % \begin{subfigure}{}
    %     \includegraphics[width=3in]{fig/acm/time_to_paper.pdf}
    %     \caption{Histogram of paper citation.}
    % \end{subfigure}
    % \caption{Dataset details of the ACM dataset.}
    \subfigure[Histogram of paper citation.]{\includegraphics[width=1.4in]{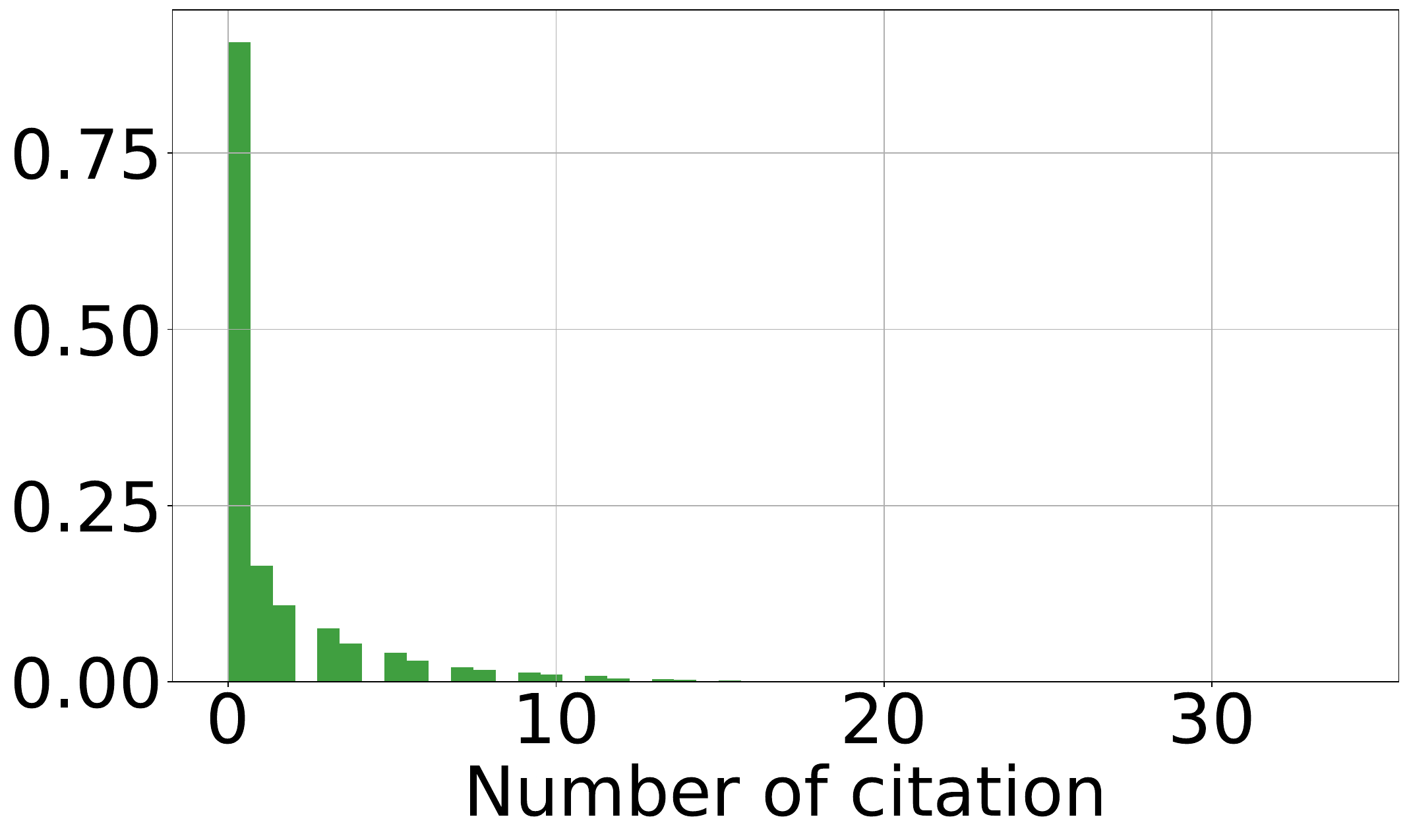}}
    \subfigure[Histogram of author publication.]{\includegraphics[width=1.4in]{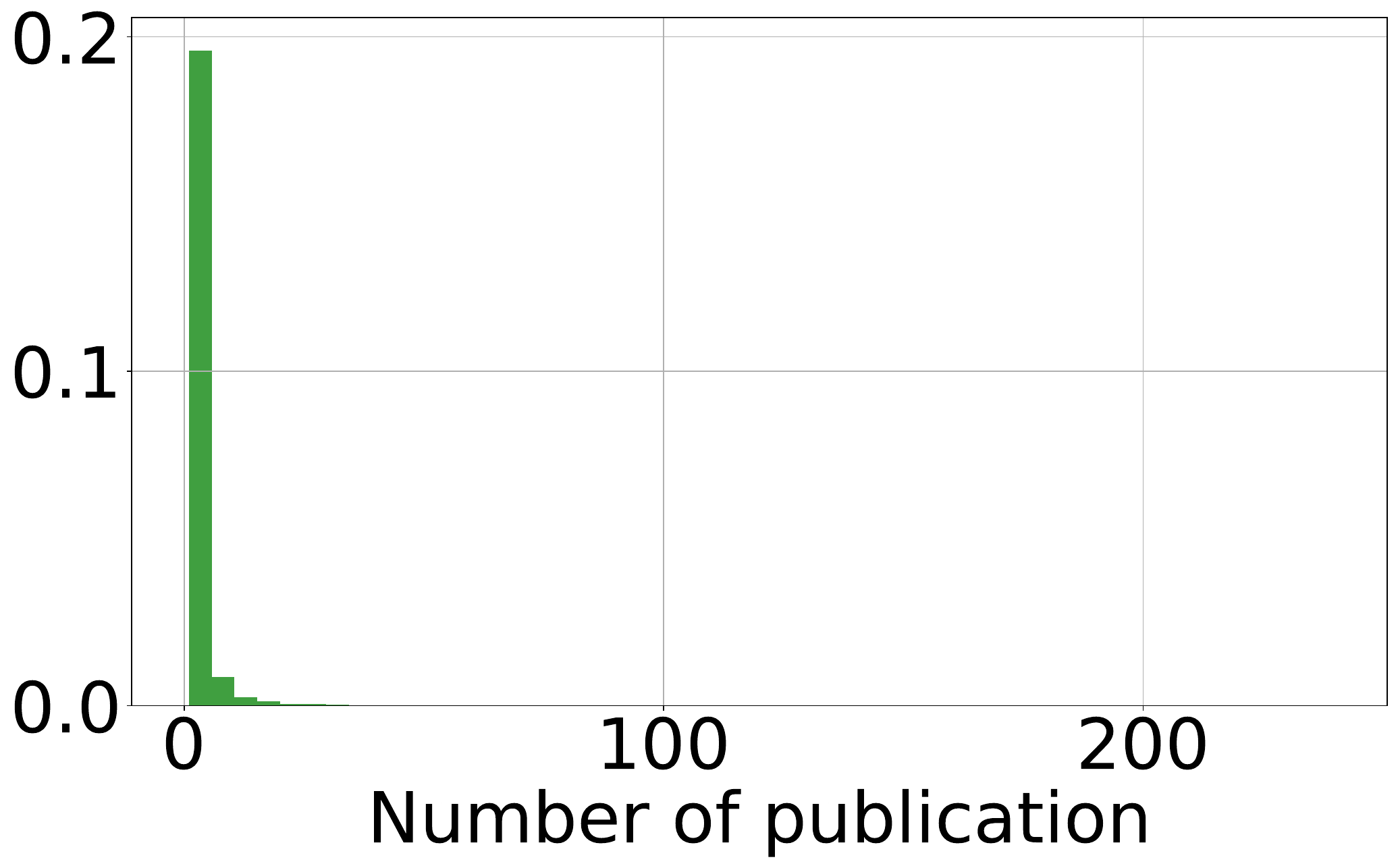}}
    
    \caption{Dataset details of the ACM dataset.}
    \label{fig:acm_dataset}
\end{figure}

\begin{figure}[h]
    \centering
    % \begin{subfigure}
    \subfigure[Papers published in each venue.]{\includegraphics[width=3in]{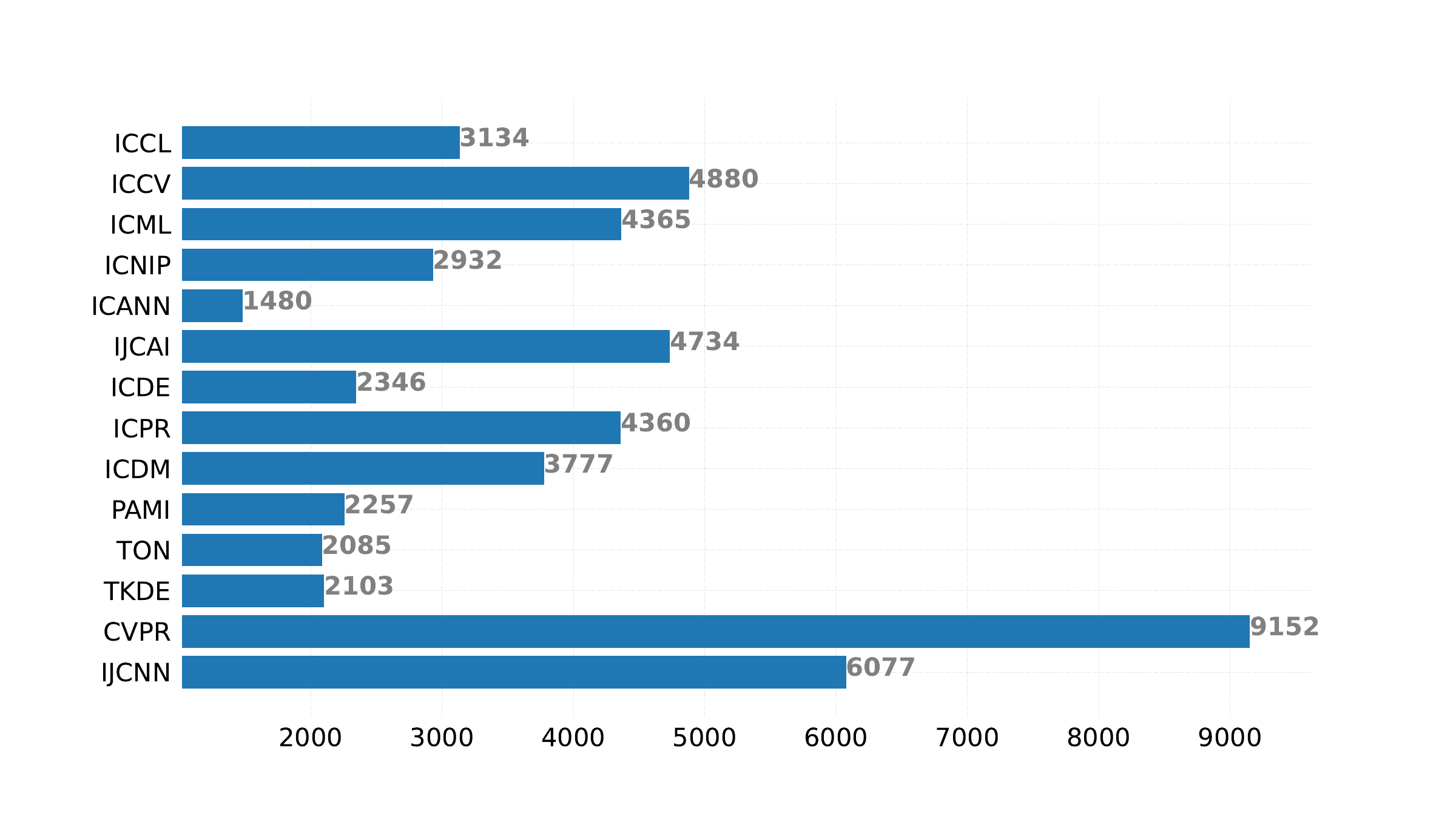}}
        % \caption{Papers published in each venue.}
    % \end{subfigure}
    
    \bigskip
    % \begin{subfigure}{}
    %     \includegraphics[width=3in]{fig/acm/time_to_paper.pdf}
    %     \caption{Histogram of paper citation.}
    % \end{subfigure}
    % \caption{Dataset details of the ACM dataset.}
    \subfigure[Papers published each yeas]{\includegraphics[width=1.4in]{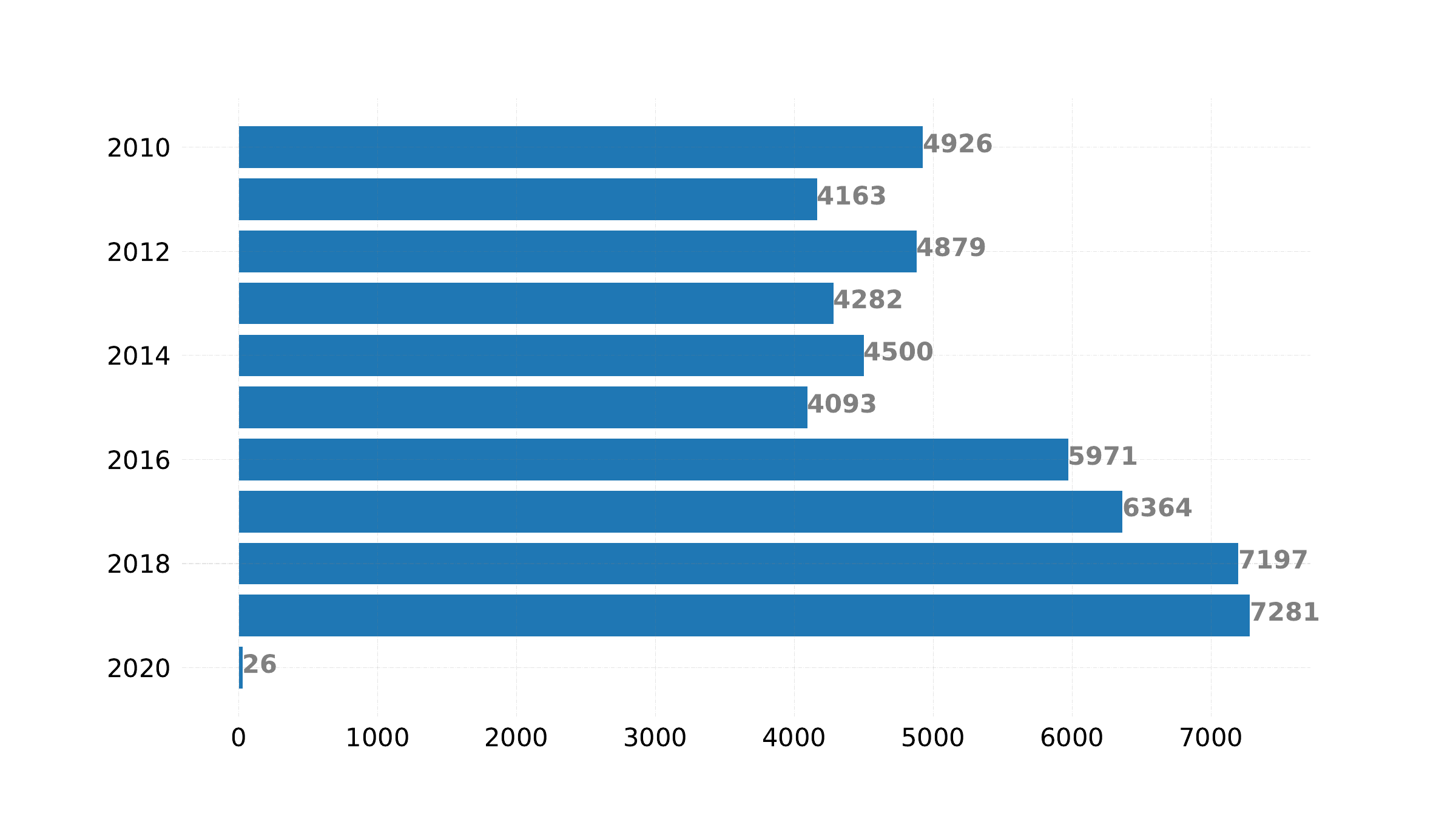}}
    \subfigure[Histogram of paper citation.]{\includegraphics[width=1.4in]{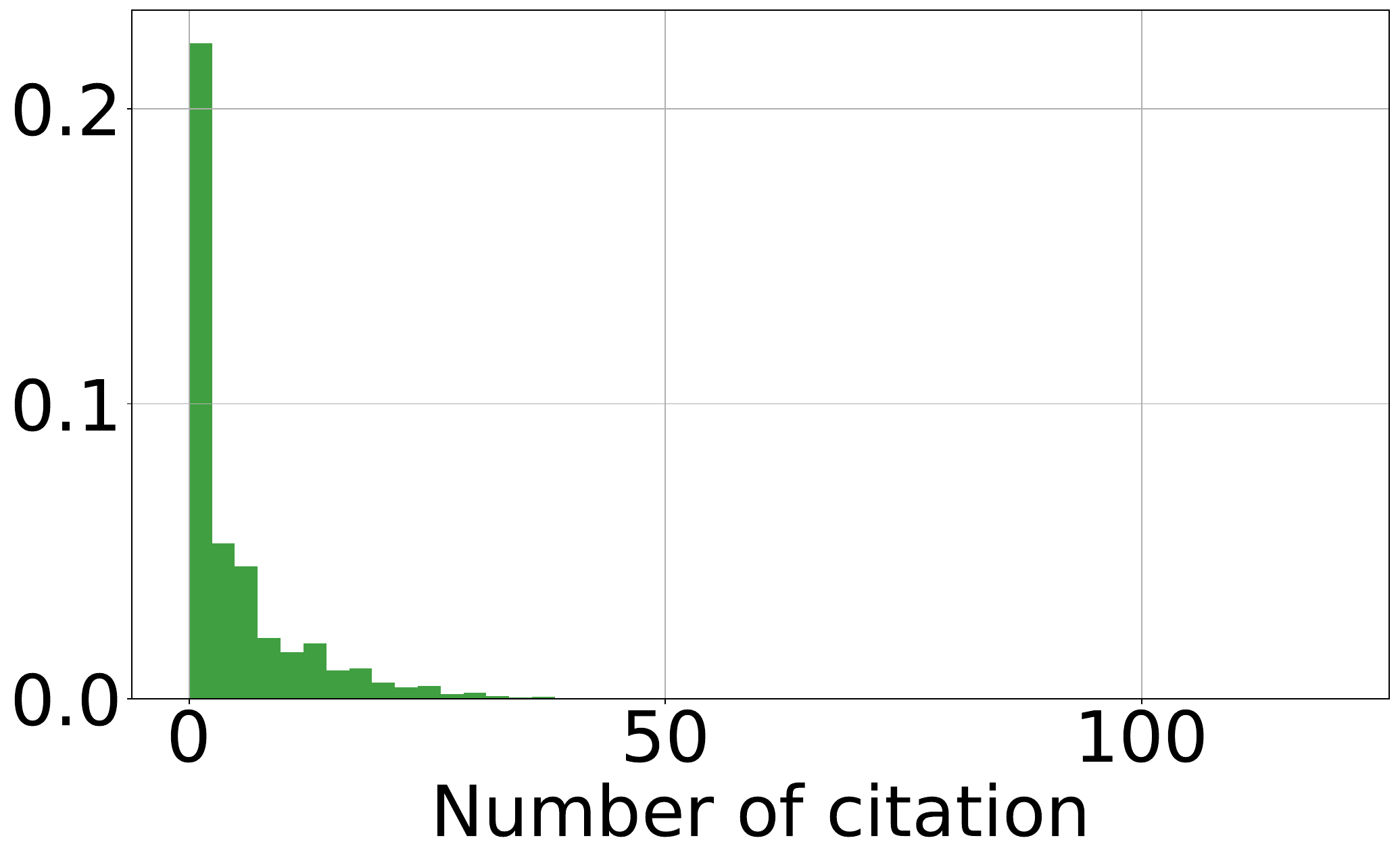}}
    
    \caption{Dataset details of the DBLP dataset.}
    \label{fig:dblp_dataset}
\end{figure}

% We applied the following pre-processing steps on all the three public datasets. 
On the ACM dataset, we select papers published in 14 venues, that is, ``WWW'', ``PAMI'', ``SIGKDD'', ``CIKM'', ``ICDM'', ``TKDE'', ``PKDD'', ``ICML'', ``PAKDD'', ``CVPR'', ``ICDE'', ``IJCAI'', ``AAAI'', ``WSDM'' from 2000 to 2016.
Similarly, on the DBLP citation network, we select papers published in between 2010 to 2020 from 14 venues including ``ICCL'', ``ICCV'', ``ICML'', ``ICNIP'', ``ICANN'', ``IJCAI'', ``ICDE'', ``ICPR'', ``ICDM'', ``PAMI'', ``TON'', ``TKDE'', ``CVPR'', and ``IJCNN''. 
The details citation information on the two citation networks are shown in Figure~\ref{fig:acm_dataset} and \ref{fig:dblp_dataset}. 
Table~\ref{table:dataset_all} shows the detailed information of the three public temporal HINs. 
We observe from the figures and table that there may be thousands of papers published in a venue. However, most of the papers have just a few citation connections with other papers, and most of the authors have published only several papers.
Thus, the degree centrality of different types of nodes is not comparable. In our work, we adopt trainable variables to scale the dynamic node degrees of different types of nodes so that their impact is well adjusted in the local aggregation process.  
Furthermore, the TREND model trains the node embeddings with the node degree prediction task, which is not suitable for a heterogeneous network with uncomparable node degrees.  

\textbf{Node feature} 
As mentioned in the paper, for citation networks, i.e., ACM and DBLP datasets, we
use the Doc2vec model to convert the
title and abstract of each paper into a real-valued vector with dimension size of 128. Moreover, we leverage the word2vec method to convert each ``Field'' into a vector of
length 128 as the feature of the ``Field'' node.
For the IMDB dataset, four kinds of movie features, i.e., release region, movie genre, and movie title, are encoded into a vector of 246 as the movie feature. To be specific, we encode the release region into a multi-hot vector of size 188, and also encode the movie genre into a multi-hot vector of 26. The movie tile is also encoded into a real-valued dense embedding of 32 with the word2vec model.
Finally, the three features are concatenated into a single feature vector of size 246.
Finally, all-zero vectors are assigned to the other node types, i.e., the ``author'' and ``venue'' nodes in citation networks, and the ``people'' node in the IMDB dataset.

\bibliographystyle{named}
\bibliography{ijcai23}
\end{document}